\documentclass[runningheads]{llncs}

\usepackage{graphicx}
\usepackage{amsmath,amssymb} % define this before the line numbering.
\usepackage{color}
\usepackage[width=122mm,left=12mm,paperwidth=146mm,height=193mm,top=12mm,paperheight=217mm]{geometry}

\usepackage{boldline}
\usepackage{multirow}
\usepackage{subcaption}
\usepackage[export]{adjustbox}
\usepackage[table]{xcolor}

\usepackage{tabularx}
\usepackage{booktabs,colortbl}
\newcolumntype{Y}{>{\centering\arraybackslash}X}
\usepackage{arydshln}
\usepackage{makecell}
\usepackage{afterpage}
\usepackage{capt-of}

\usepackage{xcolor}
\definecolor{myBlue}{HTML}{d7e7ff}
\definecolor{myRed}{HTML}{f9cfcf}
\definecolor{myGreen}{HTML}{baddb8}
\definecolor{myOrange}{HTML}{f9dfb8}

\DeclareMathOperator*{\argmin}{\arg\!\min}
\DeclareMathOperator*{\subjectto}{subject\,to\,}

\begin{document}
	\pagestyle{headings}
	\mainmatter
	
	\title{Deep Feature Factorization For Concept Discovery} % Replace with your title

	\authorrunning{E. Collins et al.}
	
	\author{Edo Collins\inst{1} \and Radhakrishna Achanta\inst{2} \and Sabine S\"usstrunk\inst{1}
	}
	\institute{School of Computer and Communication Sciences, EPFL \and
		Swiss Data Science Center, EPFL and ETHZ \\
	\email{\{edo.collins,radhakrishna.achanta,sabine.susstrunk\}@epfl.ch}}

	\maketitle
	
	\begin{abstract}
		We propose Deep Feature Factorization (DFF), a method capable of localizing similar semantic concepts within an image or a set of images. We use DFF to gain insight into a deep convolutional neural network's learned features, where we detect hierarchical cluster structures in feature space. This is visualized as heat maps, which highlight semantically matching regions across a set of images, revealing what the network `perceives' as similar. DFF can also be used to perform co-segmentation and co-localization, and we report state-of-the-art results on these tasks.
	\end{abstract}
	
	\section{Introduction}

	\begin{figure}[b]
		\centering
		\begin{tabularx}{\textwidth}{YY}
			(a) Pyramids, $k=4$ & (b) Taj Mahal, $k=3$ \\
			\multicolumn{1}{l|}{\begin{minipage}{0.49\textwidth}\includegraphics[width=\textwidth]{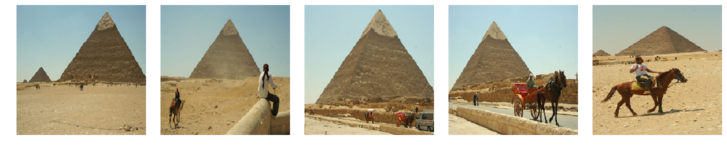}\end{minipage}~~}& \multicolumn{1}{l}{\begin{minipage}{0.49\textwidth}\includegraphics[width=\textwidth]{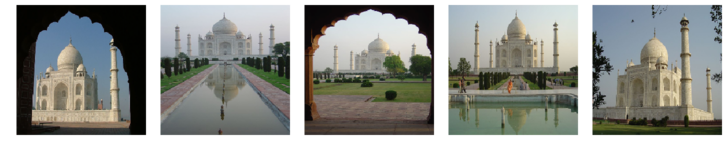}\end{minipage}} \\
			\multicolumn{1}{l|}{\begin{minipage}{0.49\textwidth}\includegraphics[width=\textwidth]{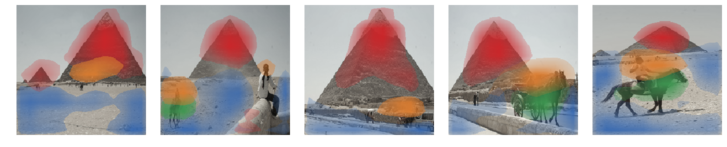}\end{minipage} } & \multicolumn{1}{l}{\begin{minipage}{0.49\textwidth}\includegraphics[width=\textwidth]{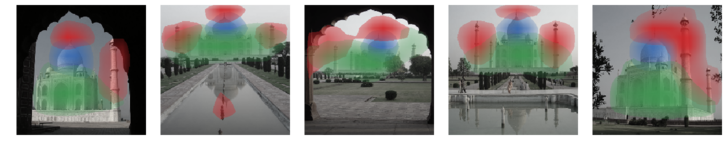}\end{minipage}}  \\

		\end{tabularx}
		
		\caption{\emph{What in this picture is the same as in the other pictures?} Our method,  Deep Feature Factorization (DFF), allows us to see how a deep CNN trained for image classification would answer this question. (a) Pyramids, animals and people correspond across images.  (b) Monument parts match with each other.  } \label{fig:icoseg1}

	\end{figure}
	
	As neural networks become ubiquitous, there is an increasing need to understand and interpret their learned representations \cite{Montavon2017,Ribeiro2016}. In the context of convolutional neural networks (CNNs), methods have been developed to explain predictions and latent activations in terms of heat maps highlighting the image regions which caused them \cite{Zhou2016,selvaraju2016grad}.
	
	In this paper, we present Deep Feature Factorization (DFF), which exploits non-negative matrix factorization (NMF) \cite{lee1999learning} applied to activations of a deep CNN layer to find semantic correspondences across images. These correspondences reflect semantic similarity as indicated by clusters in a deep CNN layer feature space. In this way, we allow the CNN to show us which image regions it `thinks' are similar or related across a set of images as well as within a single image. Given a CNN, our approach to semantic \emph{concept discovery} is unsupervised, requiring only a set of input images to produce correspondences. Unlike previous approaches \cite{Bau2017,Gonzalez2017},  we do not require annotated data to detect semantic features. We use annotated data for evaluation only.
	
	We show that when using a deep CNN trained to perform ImageNet classification \cite{ILSVRC15}, applying DFF allows us to obtain heat maps that correspond to semantic concepts. Specifically, here we use DFF to localize objects or object parts, such as the \emph{head} or \emph{torso} of an animal. We also find that parts form a hierarchy in feature space, e.g., the activations cluster for the concept \emph{body} contains a sub-cluster for \emph{limbs}, which in turn can be broken down to \emph{arms} and \emph{legs}. Interestingly, such meaningful decompositions are also found for object classes never seen before by the CNN.
	
	In addition to giving an insight into the knowledge stored in neural activations, the heat maps produced by DFF can be used to perform co-localization or co-segmentation of objects and object parts. Unlike approaches that delineate the common object across an image set, our method is also able to retrieve distinct parts \emph{within} the common object.  Since we use a pre-trained CNN to accomplish this, we refer to our method as performing weakly-supervised co-segmentation.
	
	Our main contribution is introducing Deep Feature Factorization as a method for semantic concept discovery, which can be used both to gain insight into the representations learned by a CNN, as well as to localize objects and object parts within images. We report results on several datasets and CNN architectures, showing the usefulness of our method across a variety of settings.

	\section{Related work}
	
	\subsection{Localization with CNN Activations}
	Methods for the interpretation of hidden activations of deep neural networks, and in particular of CNNs, have recently gained significant interest \cite{Montavon2017}. Similar to DFF, methods have been proposed to localize objects within an image by means of heat maps \cite{Zhou2016,selvaraju2016grad}. 
	
	In these works~\cite{Zhou2016,selvaraju2016grad}, localization is achieved by computing the importance of convolutional feature maps with respect to a particular output unit.
	These methods can therefore be seen as supervised, since the resulting heat maps are associated with a designated output unit, which corresponds to an object class from a predefined set.
	With DFF, however, heat maps are \emph{not} associated with an output unit or object class. Instead, DFF heat maps capture common activation patterns in the input, which additionally allows us to localize objects never seen before by the CNN, and for which there is no relevant output unit.
	
	\subsection{CNN Features as Part Detectors}
	The ability of DFF to localize parts stems from the CNN's ability to distinguish parts in the first place. In Gonzales et al.~\cite{Gonzalez2017} and Bau et al.~\cite{Bau2017} the authors attempt to detect learned part-detectors in CNN features, to see if such detectors emerge, even when the CNN is trained with object-level labels. They do this by measuring the overlap between feature map activations and ground truth labels from a part-level segmentation dataset. The availability of ground truth is essential to their analysis, yielding a catalog of CNN units that sufficiently correspond to labels in the dataset.

	 We confirm their observations that part detectors do indeed emerge in CNNs. However, as opposed to these  previous methods, our NMF-based approach does not rely on ground truth labels to find the parts in the input. We use labeled data for evaluation only.

	 \subsection{Non-negative Matrix Factorization}
	 Non-negative matrix factorization (NMF) has been used to analyze data from various domains, such as audio source separation~\cite{grais2011single}, document clustering \cite{xu2003document}, and face recognition \cite{guillamet2002non}.
	 
	 There has been work extending NMF to multiple layers~\cite{cichocki2006multilayer}, implementing NMF using neural networks \cite{dziugaite2015neural} and using NMF approximations as input to a neural network~\cite{vu2016combining}. However, to the best of our knowledge, the application of NMF to the activations of a pre-trained neural network, as is done in DFF, has not been previously proposed.

	\section{Method} \label{sec:Method}
	
	\begin{figure}[t]
		\centering
		\includegraphics[width=\textwidth]{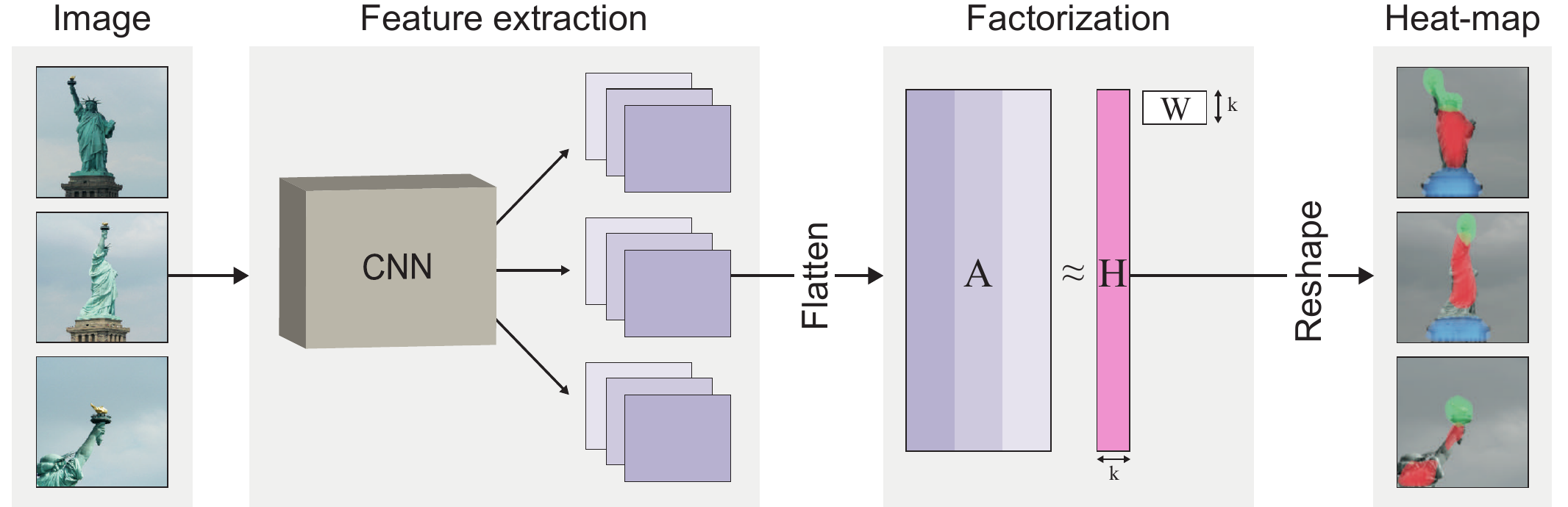}
		\caption{An illustration of Deep Feature Factorization. We extract features from a deep CNN and view them as a matrix. We apply NMF to the feature matrix and reshape the resulting $k$ factors into $k$ heat maps. See section \ref{sec:Method} for a detailed explanation. Shown: Statute of Liberty subset from iCoseg with $k=3$.} \label{fig:pipeline}
	\end{figure}
	
	\subsection{CNN Feature Space}
	In the context of CNNs, an input image $\mathcal{I}$ is seen as a tensor of dimension $h_\mathcal{I}\times w_\mathcal{I}\times c_\mathcal{I}$, where the first two dimensions are the height and the width of the image, respectively, and the third dimension is the number of color channels, e.g., 3 for RGB. Viewed this way, the first two dimensions of $\mathcal{I}$ can be seen as a spatial grid, with the last dimension being a $c_\mathcal{I}$-dimensional feature representation of a particular spatial position. For an RGB image, this feature corresponds to color.
	
	As the the image gets processed layer by layer, the hidden activation at the $\ell$th layer of the CNN is a tensor we denote $\mathcal{A}_\mathcal{I}^\ell$ of dimension $h_\ell\times w_\ell\times c_\ell$. Notice that generally $h_\ell< h_\mathcal{I},~w_\ell< w_\mathcal{I}$ due to pooling operations commonly used in CNN pipelines. The number of channels $c_\ell$ is user-defined as part of the network architecture, and in deep layers is often on the order of 256 or 512.
	
	The tensor $\mathcal{A}_\mathcal{I}^\ell$ is also called a \emph{feature map} since it has a spatial interpretation similar to that of the original image $\mathcal{I}$: the first two dimensions represent a spatial grid, where each position corresponds to a \emph{patch} of pixels in $\mathcal{I}$, and the last dimension forms a $c_\ell$-dimensional representation of the patch. The intuition behind deep learning suggests that the deeper layer $\ell$ is, the more  abstract and semantically meaningful are the $c_\ell$-dimensional features \cite{bengio2013representation}.
	
	Since a feature map represents multiple patches (depending on the size of image $\mathcal{I}$), we view them as points inhabiting the same  $c_\ell$-dimensional space, which we refer to as the CNN \emph{feature space}. Having potentially many points in that space, we can apply various methods to find directions that are `interesting'.

	\subsection{Matrix Factorization}
	Matrix factorization algorithms have been used for data interpretation for decades. For a data matrix $A$, these methods retrieve an approximation of the form:
	\begin{align} \label{eq:matrix_factorization}
	A&\approx \hat{A} = HW \\ 
	\text{s.t. } A,~\hat{A}\in &\mathcal{R}^{n\times m},~ H\in\mathcal{R}^{n\times k},~W\in\mathcal{R}^{k\times m} \nonumber
	\end{align}
	where $\hat{A}$ is a low-rank matrix of a user-defined rank $k$. A data point, i.e., a row of $A$, is explained as a weighted combination of the factors which form the rows of $W$.
	
	A classical method for dimensionality reduction is principal component analysis (PCA)~\cite{jolliffe1986principal}.
	PCA finds an optimal $k$-rank approximation (in the $\ell^2$ sense) by solving the following objective:
	\begin{equation}
	\begin{aligned}
	\text{PCA}(A, k) = & \argmin_{\hat{A}_k}
	& & \|A-\hat{A}_k \|_F^2, \\
	& \subjectto
	& & \hat{A}_k = AV_kV_k^\top,~V_k^\top V_k = I_k,
	\end{aligned}
	\end{equation}
	where $\|.\|_F$ denotes the Frobenius norm and $V_k\in\mathcal{R}^{m\times k}$. For the form of Eq. (\ref{eq:matrix_factorization}), we set $H=AV_k,~W=V_k^\top$. Note that the PCA solution generally contains negative values, which means the  combination of PCA factors (i.e., principal components) leads to the canceling out of positive and negative entries. This cancellation makes intuitive interpretation of individual factors difficult.
	
	On the other hand, when the data $A$ is non-negative, one can perform non-negative matrix factorization (NMF):
	\begin{equation}
	\begin{aligned}
	\text{NMF}(A, k) = & \argmin_{\hat{A}_k}
	& & \|A-\hat{A}_k \|_F^2, \\
	& \subjectto
	& & \hat{A}_k = HW,~ \forall ij, H_{ij} ,W_{ij} \geq 0,
	\end{aligned}
	\end{equation}
	where $H\in\mathcal{R}^{n\times k}$ and $W\in\mathcal{R}^{k\times m}$ enforce the dimensionality reduction to rank $k$. Capturing the structure in $A$ while forcing combinations of factors to be additive results in factors that lend themselves to interpretation \cite{lee1999learning}.

	\subsection{Non-negative Matrix Factorization on CNN Activations} \label{sec: NMF on CNN Activations}
	Many modern CNNs make use of the rectified linear activation function, $\max(x, 0)$, due to its desirable gradient properties. An obvious property of this function is that it results in non-negative activations. NMF is thus naturally applicable in this case.
	
	Recall the activation tensor for image $\mathcal{I}$ and layer $\ell$:
	\begin{align}
	\mathcal{A}_\mathcal{I}^\ell \in \mathbb{R}^{h\times w\times c}
	\end{align}where $\mathbb{R}$ refers to the set of non-negative real numbers. To apply matrix factorization, we partially flatten $\mathcal{A}$ into a matrix whose first dimension is the product of $h$ and $w$:
	\begin{align}
	A_\mathcal{I}^\ell \in \mathbb{R}^{(h\cdot w)\times c}
	\end{align}
	
	Note that the matrix $A_\mathcal{I}^\ell$ is effectively a `bag of features' in the sense that the spatial arrangement has been lost, i.e., the rows of $A_\mathcal{I}^\ell$ can be permuted without affecting the result of factorization. We can naturally extend factorization to a set of $n$ images, by vertically concatenating their features together:
	\begin{align}
	A = \begin{bmatrix}
	A_1^\ell\\[0.3em]
	\vdots \\[0.3em]
	A_n^\ell 
	\end{bmatrix}
	\in \mathbb{R}^{(n\cdot h\cdot w)\times c}
	\end{align}
	
    For ease of notation we assumed all images are of equal size, however, there is no such limitation as images in the set may be of any size.
	By applying NMF to $A$ we obtain the two matrices from Eq. \ref{eq:matrix_factorization}, $H\in \mathbb{R}^{(n\cdot h\cdot w)\times k}$ and $W\in \mathbb{R}^{k\times c}$.
	
	\subsection{Interpreting NMF Factors} \label{sec:Interpreting NMF Factors}
	
	The result returned by the NMF consists of $k$ \emph{factors}, which we will call DFF factors, where $k$ is the predefined rank of the approximation. %The matrices $W$ and $H$ both consist of $k$ columns.
	
	\subsubsection{The $W$ Matrix}
	Each row $W_j$ ($1\leq j\leq k$) forms a $c$-dimensional vector in the CNN feature space. Since NMF can be seen as performing clustering \cite{ding2005equivalence}, we view a factor $W_j$ as a centroid of an activation cluster, which we show corresponds to coherent object or object-part.
	
	\subsubsection{The $H$ Matrix}
	The matrix $H$ has as many rows as the activation matrix $A$, one corresponding to every spatial position in every image. Each row $H_i$ holds coefficients for the weighted sum of the $k$ factors in $W$, to best approximate the $c$-dimensional $A_i$.
	
	Each column $H_j$ ($1\leq j\leq k$) can be reshaped into $n$ \textbf{heat maps} of dimension $h\times w$, which highlight regions in each image that correspond to the factor $W_j$. These heat maps have the same spatial dimensions as the CNN layer which produced the activations, often low. To match the size of the heat map with the input image, we upsample it with bilinear interpolation.

	\section{Experiments} \label{sec:experiments}
	
	In this section we first show that DFF can produce a hierarchical decomposition into semantic parts, even for sets of very few images (section \ref{sec:icoseg_eval}). We then move on to larger-scale, realistic datasets where we show that DFF can perform state-of-the-art weakly-supervised object co-localization and co-segmentation, in addition to part co-segmentation (sections \ref{sec:pascal2007} and \ref{sec:pascal_parts}).

	\subsection{Implementation Details}

	\subsubsection{NMF.} NMF optimization with multiplicative updates \cite{lee2001algorithms} relies on dense matrix multiplications, and can thus benefit from fast GPU operations. Using an NVIDIA Titan X, our implementation of NMF can process over 6K images of size $224\times 224$ at once with $k=5$,  and requires less than a millisecond per image. Our code is available online.

	\subsubsection{Neural Network Models.}
	We consider five network architectures in our experiments, namely VGG-16 and VGG-19 \cite{Simonyan2014}, with and without batch-normalization \cite{Joffe2015}, as well as ResNet-101 \cite{He2016}. We use the publicly available models from \cite{Paszke2017}.

	\subsection{Segmentation and localization methods} \label{sec:baseline}
	In addition to gaining insights into CNN feature space, DFF has utility for various tasks with subtle but important differences in naming:
	\begin{itemize}
		\item \textbf{Segmentation vs. Localization} is the difference between predicting pixel-wise binary masks and predicting bounding boxes, respectively.
		\item \textbf{Segmentation vs. co-segmentation} is the distinction between segmenting a single image into regions and jointly segmenting multiple images, thereby producing a correspondence between regions in different images (e.g., \emph{cat}s in all images belong to the same segment).
		\item \textbf{Object co-segmentation vs. Part co-segmentation}. Given a set of images representing a common object, the former performs binary background-foreground separation where the foreground segment encompasses the entirety of the common object (e.g., \emph{cat}). The latter, however, produces $k$ segments, each corresponding to a \emph{part} of the common object (e.g., \emph{cat head}, \emph{cat legs}, etc.).
	\end{itemize}
	
	When applying DFF with $k=1$ can we compare our results against object co-segmentation (background-foreground separation) methods and object co-localization methods.
	
	In section \ref{sec:icoseg} we compare DFF against three state-of-the-art co-segmentation methods.	 
	The supervised method of Vicente et al.~\cite{vicente2011object} chooses among multiple segmentation proposals per image by learning a regressor to predict, for pairs of images, the overlap between their proposals and the ground truth. Input to the regressor included per-image features, as well as pairwise features.	 
	The methods Rubio et al.~\cite{rubio2012unsupervised} and Rubinstein et al.~\cite{Rubinstein13Unsupervised} are unsupervised and rely on a Markov random field formulation, where the unary features are based on surface image features and various saliency heuristics. For pairwise terms, the former method uses a per-image segmentation into regions, followed by region-matching across images.  The latter approach uses a dense pairwise correspondence term between images based on local image gradients.
	
	In section \ref{sec:pascal2007} we compare against several state-of-the-art object co-localization methods.
	Most of these methods operate by selecting the best of a set of object proposals, produced by a pre-trained CNN \cite{li2016image} or an object-saliency heuristic \cite{cho2015,joulin2014}.
	The authors of \cite{le2017co} present a method for unsupervised object co-localization that, like ours, also makes use of CNN activations. Their approach is to apply $k$-means clustering to globally max-pooled activations, with the intent of clustering all highly active CNN filters together. Their method therefore produces a \emph{single} heat map, which is appropriate for object co-segmentation, but \emph{cannot} be extended to part co-segmentation.
	
	When  $k>1$, we use DFF to perform part co-segmentation. Since we have not come across examples of part co-segmentation in the literature, we compare against a method for supervised part segmentation, namely Wang et al. \cite{wang2015semantic} (Table \ref{tab:wang} in section \ref{sec:pascal_parts}).  Their method relies on a compositional model with strong explicit priors w.r.t to part size, hierarchy and symmetry. We also show results for two baseline methods described in~\cite{wang2015semantic}: PartBB+ObjSeg where segmentation masks are produced by intersecting part-bounding-boxes~\cite{Chen2014} with whole-object segmentation masks~\cite{BharathECCV2014}. The method PartMask+ObjSeg is similar, but here bounding-boxes are replaced with the best of 10 pre-learned part masks.

	\subsection{Experiments on iCoseg} \label{sec:icoseg}
	
	\subsubsection{Dataset}
	The iCoseg dataset \cite{batra2010icoseg} is a popular benchmark for co-segmentation methods. As such, it consists of 38 sets of images, where each image is annotated with a pixel-wise mask encompassing the main object common to the set.
	Images within a set are uniform in that they were all taken on a single occasion, depicting the same objects. The challenging aspect of this datasets lies in the significant variability with respect to viewpoint, illumination, and object deformation.
	
	We chose five sets and further labeled them with pixel-wise object-part masks (see Table \ref{tab:icoseg}). This process involved partitioning the given ground truth mask into sub-parts.  We also annotated common background objects, e.g., \emph{camel} in the \emph{Pyramids} set (see Figure \ref{fig:icoseg1}). Our part-annotation for iCoseg is available online. The number of images in these sets ranges from as few as 5 up to 41. When comparing against \cite{vicente2011object} and \cite{rubio2012unsupervised} in Table \ref{tab:icoseg}, we used the subset of iCoseg used in those papers.

	\subsubsection{Part co-segmentation} \label{sec:icoseg_eval}
\begin{figure}[t]
		\setlength{\tabcolsep}{0.8pt}
		\begin{tabular}{lcc}
						& (a) Gymnastics1 & (b) Elephants \\
			& \multicolumn{1}{l|}{\begin{minipage}{0.48\textwidth}\includegraphics[width=\textwidth]{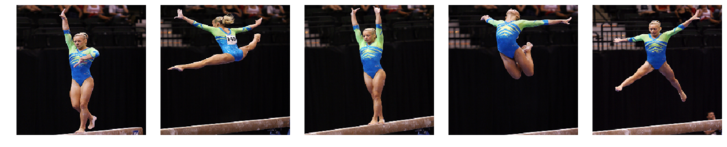}\end{minipage} }& \multicolumn{1}{l}{\begin{minipage}{0.48\textwidth}\includegraphics[width=\textwidth]{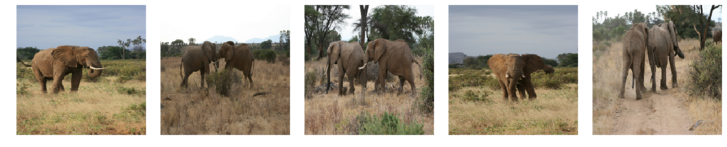}\end{minipage}} \\
			
			\rotatebox[origin=c]{90}{\footnotesize$k\!=\!1$}& \multicolumn{1}{l|}{\begin{minipage}{0.48\textwidth}\includegraphics[width=\textwidth]{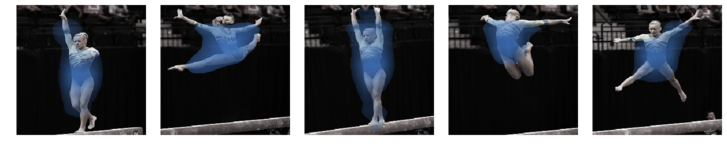}\end{minipage} } & \multicolumn{1}{l}{\begin{minipage}{0.48\textwidth}\includegraphics[width=\textwidth]{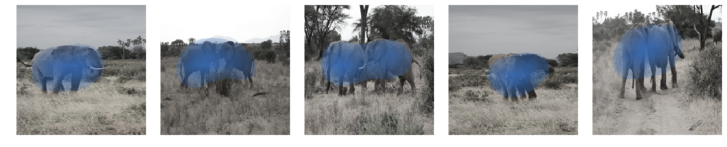}\end{minipage}}  \\
			
			\rotatebox[origin=c]{90}{\footnotesize$k\!=\!2$}& \multicolumn{1}{l|}{\begin{minipage}{0.48\textwidth}\includegraphics[width=\textwidth]{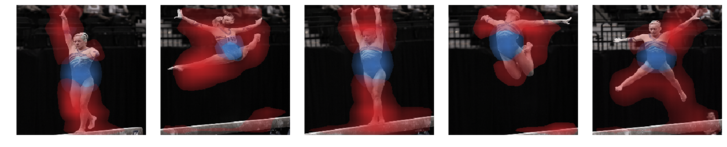}\end{minipage} } & \multicolumn{1}{l}{\begin{minipage}{0.48\textwidth}\includegraphics[width=\textwidth]{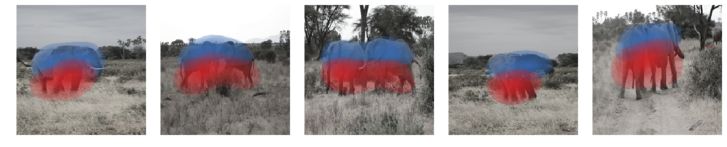}\end{minipage} } \\
			
			\rotatebox[origin=c]{90}{\footnotesize$k\!=\!3$}& \multicolumn{1}{l|}{\begin{minipage}{0.48\textwidth}\includegraphics[width=\textwidth]{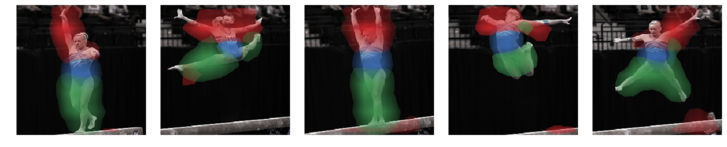}\end{minipage}~~}  & \multicolumn{1}{l}{\begin{minipage}{0.48\textwidth}\includegraphics[width=\textwidth]{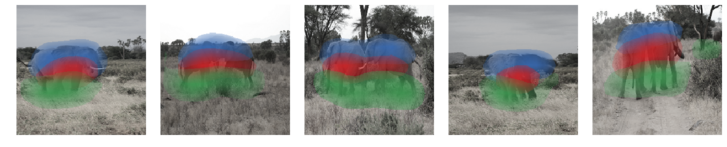}\end{minipage} } \\
			
			\rotatebox[origin=c]{90}{\footnotesize$k\!=\!4$}\normalsize& \multicolumn{1}{l|}{\begin{minipage}{0.48\textwidth}\includegraphics[width=\textwidth]{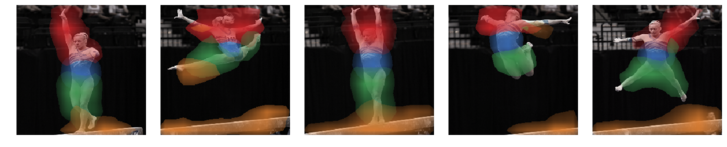}\end{minipage}}  & \multicolumn{1}{l}{\begin{minipage}{0.48\textwidth}\includegraphics[width=\textwidth]{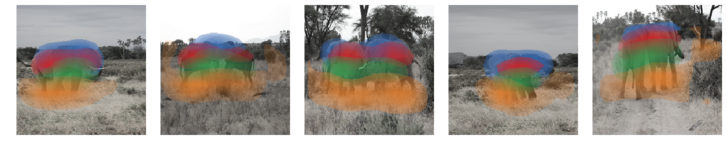}\end{minipage}}  \\

		\end{tabular}
		
		\caption{Example DFF heat maps for images of two sets from iCoseg. Each row shows a separate factorization where the number of DFF factors $k$ is incremented. Different colors correspond to the heat maps of the $k$ different factors. DFF factors correspond well to distinct object parts. This Figure visualizes the data in Table \ref{tab:icoseg}, where heat map color corresponds with row color. (Best viewed electronically with a color display)}  \label{fig:icoseg2}
		
	\end{figure}

	For each set in iCoseg, we obtained activations from the deepest convolutional layer of VGG19 ({\tt conv5\_4}), and applied NMF to these activations with increasing values of $k$. The resulting heat maps can be seen in Figures \ref{fig:icoseg1} and \ref{fig:icoseg2}.
	
	Qualitatively, we see a clear correspondence between DFF factors and coherent object-parts, however, the heat maps are coarse. Due to the low resolution of deep CNN activations, and hence of the heat map, we get blobs that do not perfectly align with the underlying region of interest. We therefore also report additional results with a post-processing step to refine the heat maps, described below.
	
	We notice that when $k=1$, the single DFF factor corresponds to a whole object, encompassing multiple object-parts. This, however, is not guaranteed, since it is possible that for a set of images, setting $k=1$ will highlight the \emph{background} rather than the foreground. 
	Nonetheless, as we increase $k$, we get a decomposition of the object or scene into individual parts. This behavior reveals a hierarchical structure in the clusters formed in CNN feature space. 
	
	For instance, in Figure \ref{fig:icoseg2} (a), we can see that $k=1$ encompasses most of gymnast's body, $k=2$ distinguished her midsection from her limbs, $k=3$ adds a finer distinctions between arms and legs, and finally $k=4$ adds a new component that localizes the beam. This observation also indicates the CNN has learned representation that `explains' these concepts with invariance to pose, e.g., leg positions in the 2nd, 3rd, and 4th columns.
	
	A similar decomposition into legs, torso, back, and head can be seen for the elephants in Figure \ref{fig:icoseg2} (b). This shows that we can localize different objects and parts even when they are all common across the image set. Interestingly, the decompositions shown in Figure \ref{fig:icoseg1} exhibit similar high semantic quality in spite of their dissimilarity to the ImageNet training data, as neither pyramids nor the Taj Mahal are included as class labels in that dataset.
	We also note that as some of the given sets contain as few as 5 images (Figure \ref{fig:icoseg1} (b) comprises the whole set), our method does not require many images to find meaningful structure.

	\subsubsection{Object and Part co-segmentation}
	We operationalize DFF to perform co-segmentation. To do so we have to first annotate the factors as corresponding to specific ground-truth parts. This can be done manually (as in Table \ref{tab:wang}) or automatically given ground truth, as described below. We report the intersection-over-union ($IoU$) score of each factor with its associated parts in Table \ref{tab:icoseg}.
	
	Since the heat maps are of low-resolution, we refine them with post processing. We define a dense conditional random field (CRF) over the heat maps. We use the filter-based mean field approximate inference~\cite{krahenbuhl2011efficient}, where we employ guided filtering \cite{he2013guided} for the pairwise term, and use the biliniearly upsampled DFF heat maps as unary terms. We refer to DFF with post-processing 'DFF-CRF'.
	
	Each heat map is converted to a binary mask using a thresholding procedure. For a specific DFF factor $f$ ($1\leq f\leq k$), let $\{H(f, 1),\cdots, H(f,n)\}$ be the set of $n$ heat maps associated with $n$ input images, The value of a pixel in the binary map $B(f,i)$ of factor $f$ and image $i$ is 0 if its intensity is lower than the 75th percentile of entries in the set of heat maps $\{H(f,j) | 1\leq j\leq n\}$.
	
	We associate parts with factors by considering how well a part is covered by a factor's binary masks. We define the \emph{coverage} of part $p$ by factor $f$ as:
	\begin{align}
	Cov_{f,p} = \frac{|\sum_i B(f,i) \bigcap P(p,i)|}{|\sum_i P(p,i)|}
	\end{align}
	The coverage is the percentage of pixels belonging to $p$ that are set to 1 in the binary maps$\{B(f,i) | 1\leq i\leq n\}$. We associate the part $p$ with factor $f$ when  $Cov_{f,p}>Cov_{\text{th}}$. We experimentally set the threshold $Cov_{\text{th}}=0.5$.
	
	Finally, we measure the $IoU$ between a DFF factor $f$ and its $m$ associated ground-truth parts $\{p^{(f)}_1,\cdots,p^{(f)}_m\}$ similarly to \cite{Bau2017}, specifically by considering the dataset-wide $IoU$ :
	\begin{align}
	&P_f(i) = \bigcup_j^m P(p^{(f)}_j) \\
	&IoU_{f,p} = \frac{|\sum_i B_i \bigcap P_f(i)|}{|\sum_i B_i \bigcup P_f(i)|}
	\end{align}

	\begin{table}[t] 
	\fontsize{6pt}{4.8pt} \selectfont
	\renewcommand{\arraystretch}{2}
	\centering
	\begin{tabular}{c|l@{\hspace{1pt}}l:l@{\hspace{1pt}}l:l@{\hspace{1pt}}l:l@{\hspace{1pt}}l:l@{\hspace{1pt}}l} \hlineB{2}
		
		Method&  \multicolumn{2}{c}{Elephants} & \multicolumn{2}{:c}{Taj Mahal} & \multicolumn{2}{:c}{Pyramids} & \multicolumn{2}{:c}{Gymnastics1} & \multicolumn{2}{:c}{Statue of Liberty} \\ \hline
		\multicolumn{11}{c}{\textbf{\tiny Object co-segmentation}} \\ \hline
		
		\multirow{1}{*}{Vicente \cite{vicente2011object}} & \emph{whole} & 43 & \emph{whole} & \textbf{91} &\multicolumn{2}{:c}{-} & \multicolumn{2}{:c}{-} & \emph{whole} & \textbf{94}\\
		
		\multirow{1}{*}{Rubio \cite{rubio2012unsupervised}} & \emph{whole} & 75 & \emph{whole} & 89 &\multicolumn{2}{:c}{-} & \multicolumn{2}{:c}{-} & \emph{whole} & 92\\
		
		\multirow{1}{*}{Rubinstein \cite{Rubinstein13Unsupervised}} & \emph{whole} & 63 & \emph{whole} & 48 & \emph{whole} & 57 & \emph{whole} & \textbf{94} & \emph{whole} & 70\\

		\multirow{1}{*}{DFF, $k$=1} & \cellcolor{myBlue}\emph{whole} & \cellcolor{myBlue}65 & \emph{whole} & 41 & \emph{whole} & 57 & \cellcolor{myBlue}\emph{whole} & \cellcolor{myBlue}43 & \emph{whole} & 49\\ 
		\multirow{1}{*}{DFF-CRF, $k$=1} & \emph{whole} & \textbf{76} & \emph{whole} & 51 & \emph{whole} & \textbf{70} & \emph{whole} & 52 & \emph{whole} & 62\\ \hline
		
		 \multicolumn{11}{c}{\textbf{\tiny Part co-segmentation}} \\ \hline
		 & \cellcolor{myBlue}\emph{torso/back/head} & \cellcolor{myBlue}59 & \emph{dome} & 33 & \emph{animal} & 36 & \cellcolor{myBlue}\emph{torso/waist} & \cellcolor{myBlue}35 & \emph{torso} & 36\\
		\multirow{-2}{*}{DFF, $k$=2} & \cellcolor{myRed}\emph{torso/leg} & \cellcolor{myRed}35 & \emph{tower/building} & 46 & \emph{pyramid} & 56 & \cellcolor{myRed} \emph{arm/leg/head} & \cellcolor{myRed}20 & \emph{torch/base/head} & 28\\\hline
		\rowcolor{myBlue}
		\cellcolor{white}
		& \emph{back/head} & 46 & \emph{building} & 45 & 	\cellcolor{white} \emph{background} & 	\cellcolor{white} 27 & \emph{torso/waist} & 38 & \emph{base} & 14\\
		\rowcolor{myRed}
		\cellcolor{white}
		&  \emph{torso} & 25 & \emph{dome} & 40 & 	\cellcolor{white} \emph{pyramid} & 	\cellcolor{white} 55 & \emph{arm/head} & 22 & \emph{torso} & 39\\
		\rowcolor{myGreen}
		\cellcolor{white}
		\multirow{-3}{*}{DFF, $k$=3} & \emph{leg} & 21 & \emph{tower} & 13 & 	\cellcolor{white} \emph{animal} & 	\cellcolor{white} 36 & \emph{leg} & 33 & \emph{torch/head} & 23\\\hline

		& \cellcolor{myBlue}\emph{torso/back/head} & \cellcolor{myBlue}58  & \emph{building} & 72 & \cellcolor{myBlue}\emph{background} &  \cellcolor{myBlue}27 & \cellcolor{myBlue}\emph{torso/waist} & \cellcolor{myBlue}40 & \emph{torso} & 39\\

		& \cellcolor{myRed}\emph{head} & \cellcolor{myRed}36 & \emph{dome} & 43 & \cellcolor{myRed}\emph{pyramid} & \cellcolor{myRed}52 & \cellcolor{myRed}\emph{torso/arm/head} & \cellcolor{myRed}33 & \emph{background} & 44\\

		& \cellcolor{myGreen}\emph{torso} & \cellcolor{myGreen}20 & \emph{background} & 08 & \cellcolor{myGreen}\emph{animal} & \cellcolor{myGreen}37 & \cellcolor{myGreen}\emph{leg} & \cellcolor{myGreen}37 & \emph{torch/head} & 26\\

		\multirow{-4}{*}{DFF, $k$=4} & \cellcolor{myOrange}\emph{leg} & \cellcolor{myOrange}16  & \emph{tower} & 16 & \cellcolor{myOrange}\emph{person} & \cellcolor{myOrange}12 & \cellcolor{myOrange}\emph{background} & \cellcolor{myOrange}14 & \emph{base} & 40\\\hline
		\hlineB{2} 
	\end{tabular}\\[3ex]
	
	\caption{Object and part discovery and segmentation on five iCoseg image sets. Part-labels are automatically assigned to DFF factors, and are shown with their corresponding $IoU$-scores. Our results show that clusters in CNN feature space correspond to coherent parts. More so, they indicate the presence of a cluster hierarchy in CNN feature space, where part-clusters can be seen as sub-clusters within object-clusters (See Figures \ref{fig:icoseg1}, \ref{fig:pipeline} and \ref{fig:icoseg2} for visual comparison. Row color corresponds with heat map color). With $k=1$, DFF can be used to perform object co-segmentation, which we compare against state-of-the-art methods. With $k>1$ DFF can be used to perform part co-segmentation, which current co-segmentation methods are not able to do.
	 } \label{tab:icoseg}
\end{table}

	In the top of Table \ref{tab:icoseg} we report results for object co-segmentation ($k=1$) and show that our method is comparable with the supervised approach of \cite{vicente2011object} and domain-specific methods of \cite{rubio2012unsupervised} and \cite{Rubinstein13Unsupervised}.

	The bottom of Table \ref{tab:icoseg} shows the labels and $IoU$-scores for part co-segmentation on the five image sets of iCoseg that we have annotated. These scores correspond to the visualizations of Figures \ref{fig:icoseg1} and \ref{fig:icoseg2} and confirm what we observe qualitatively.
	
\begin{figure}[t]
		\centering
		\begin{tabular}{cccc}
			\rotatebox[origin=c]{90}{\centering Avg. $IoU$} \hspace{1.0pt} &
			\raisebox{-.5\height}{\includegraphics[width=.45\textwidth]{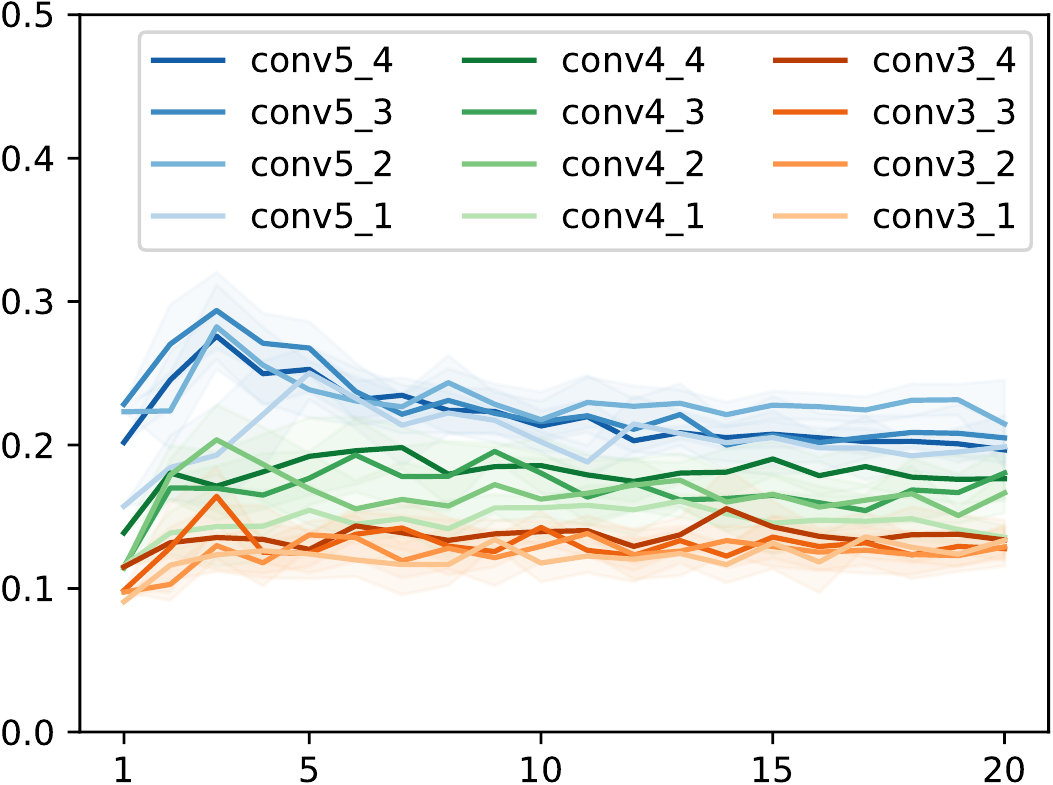}} &
			\hspace{1cm}&
			\raisebox{-.5\height}{\includegraphics[width=.45\textwidth]{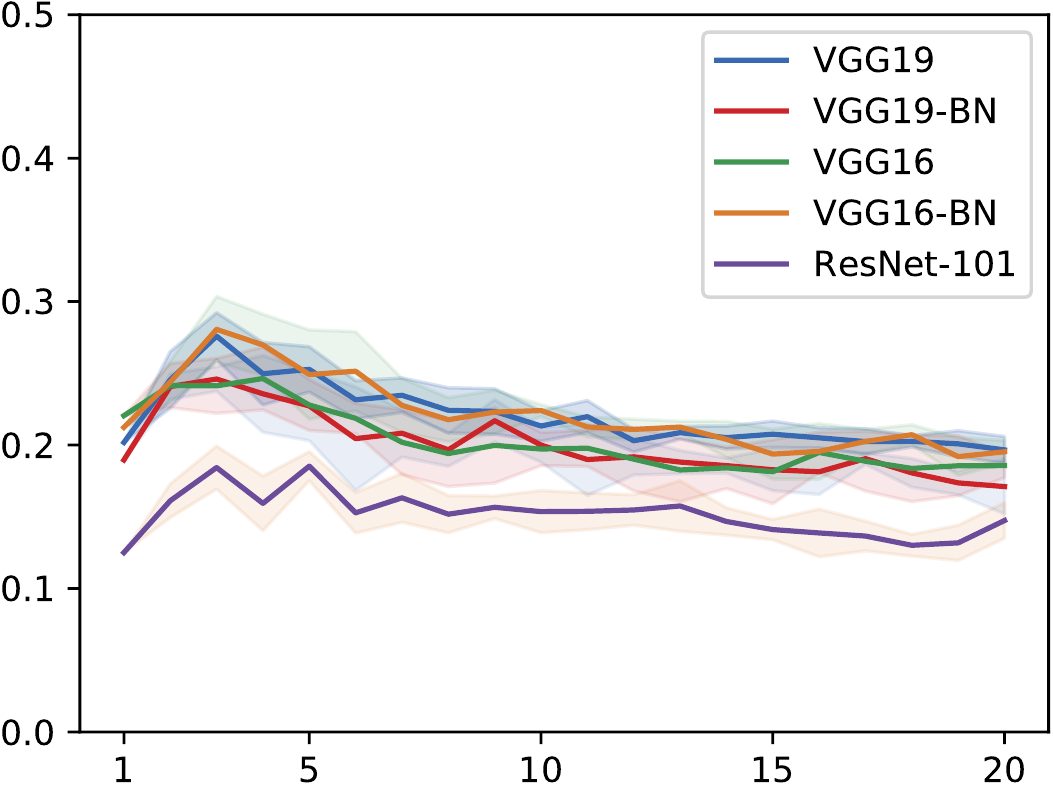}} \\
			& \multicolumn{3}{c}{Number of factors $k$} \\
			& (a) & & (b)
		\end{tabular}

		\caption{(a) Average $IoU$ score for DFF on iCoseg. for (a) different VGG19 layers and (b) the deepest convolutional layer for other CNN architectures. Expectedly, different convolutional blocks show a clear difference in matching up with semantic parts, as CNN features capture more semantic concepts. The optimal value for $k$ is data dependent but is usually below 5. We see also that DFF performance is relatively uniform for the VGG family of models.} \label{fig:icoseg_coh}
	\end{figure}

	We can characterize the quality of a factorization as the average $IoU$ of each factor with its single best matching part (which is not the background). In Figure \ref{fig:icoseg_coh} (a) we show the average  $IoU$ for different layer of VGG-19 on iCoseg as the value of $k$ increases. The variance shown is due to repeated trials with different NMF initializations. There is a clear gap between convolutional blocks. Performance with in a block does not strictly follow the linear order of layers.
	
	We also see that the optimal value for $k$ is between 3 and 5. While this naturally varies for different networks, layers, and data batches, another deciding factor is the resolution of the part ground truth. As $k$ increases, DFF heat maps become more localized, highlighting regions that are beyond the granularity of the ground truth annotation, e.g., a pair of factors that separates \emph{leg} into \emph{ankle} and \emph{thigh}.
	In Figure \ref{fig:icoseg_coh} (b) we show that DFF performs similarly within the VGG family of models. For ResNet-101 however, the average $IoU$ is distinctly lower.

	\subsection{Object Co-Localization on PASCAL VOC 2007} \label{sec:pascal2007}

	\subsubsection{Dataset}
	PASCAL VOC 2007 has been commonly used to evaluate whole object co-localization methods. Images in this dataset often comprise several objects of multiple classes from various viewpoints, making it a challenging benchmark. As in previous work \cite{le2017co,cho2015,joulin2014}, we use the \emph{trainval} set for evaluation and filter out images that only contain objects which are marked as \emph{difficult} or \emph{truncated}. The final set has 20 image sets (one per class), with 69 to 2008 images each.
	
	\subsubsection{Evaluation}
	The task of co-localization involves fitting a bounding box around the common object in a set of image.
	With $k=1$, we expect DFF to retrieve a heat map which localizes that object. 
	
	As described in the previous section, after optionally filtering DFF heat maps using a CRF, we convert the heat maps to binary segmentation masks. We follow \cite{selvaraju2016grad} and extract a single bounding box per heat map by fitting a box around the largest connected component in the binary map. 
	
	We report the standard CorLoc score \cite{deselaers2012weakly} of our localization. The CorLoc score is defined as the percentage of predicted bounding boxes for which there exists a matching ground truth bounding box. Two bounding boxes are deemed matching if their $IoU$ score exceeds 0.5.
	
	The results of our method are shown in Table \ref{tab:P07-coloc}, along with previous methods (described in section \ref{sec:baseline}). Our method compares favorably against previous approaches. For instance, we improve co-localization for the class \emph{dog} by 16\% higher CorLoc and achieve better co-localization on average, in spite of our approach being simpler and more general.

		\begin{table} 
		\tiny
		\centering
		\begin{tabularx}{\textwidth}{X|cccccccccccccccccccc|c} \hlineB{2}
			Method & aero & bicy & bird & boa & bot & bus & car & cat & cha & cow & dtab & dog & hors & mbik & pers & plnt & she & sofa & trai & tv & mean \\ \hline
			Joulin~\cite{joulin2014} & 33 & 17 & 21 & 18 & 5 & 27 & 33 & 41 & 6 & 29 & \textbf{35} & 32 & 26 & 40 & 18 & 12 & 25 & 28 & 36 & 12 & 25.60 \\
			Cho~\cite{cho2015} & 50 & 43 & 30 & 19 & 4 & 62 & \textbf{65} & 43 & \textbf{9} & 49 & 12 & 44 & 64 & 57 & 15 & 9 & 31 & 34 & 62 & \textbf{32} & 36.60 \\
			Li~\cite{li2016image} & \textbf{73} & 45 & 43 & 28 & 7 & 53 & 58 & 45 & 6 & 48 & 14 & 47 & \textbf{69} & 67 & \textbf{24} & 13 & \textbf{52} & 26 & 65 & 17 & 40.00 \\
			Le (A)~\cite{le2017co} & 70 & 52 & 44 &  \textbf{30} & 5 & 56 & 60 & 59 & 6 & 49 & 16 & 51 & 59 & 67 & 23 & 12 & 47 & 27 & 59 & 16 & 40.36 \\
			Le( V)~\cite{le2017co} & 72 & \textbf{62} & 48 & 28 & \textbf{12} & \textbf{64} & 59 & 72 & 6 & 37 & 12 & 45 & 67 & 72 & 19 & 11 & 37 & 29 & \textbf{67} & 23 & 41.97 \\ \hline
			\textbf{DFF}  & 61 & 49 & \textbf{54} & 20 & 10 & 60 & 46 & \textbf{79} & 4 & 51 & 32 & \textbf{67} & 66 & 70 & 19 & \textbf{15} & 40 & 32 & 66 & 20 & 42.87 \\ 
	
			\textbf{DFF-CRF} & 64 & 47 & 50 & 16 & 10 & 62 & 52 & 75 & 8 & \textbf{53} &  \textbf{35} & 65 & 65 & \textbf{72} & 16 & 14 & 41 & \textbf{36} & 63 & 30 & \textbf{43.51}  \\
			
			\hlineB{2}  
		\end{tabularx}\\[1ex]
		\caption{Co-localization results for PASCAL VOC 2007 with DFF $k=1$. Numbers indicate CorLoc scores. Overall, we exceed the state-of-the-art approaches using a much simpler method.} \label{tab:P07-coloc}
	\end{table}

	\subsection{Part Co-segmentation in PASCAL-Parts} \label{sec:pascal_parts}

	\begin{table}[t]
		\centering
		\small
		\setlength{\tabcolsep}{1mm}
		\begin{tabularx}{\textwidth}{Xccc:ccc} \hline
			\multicolumn{1}{c}{\multirow{2}{*}{Method}} & \multicolumn{3}{c}{\emph{cow}} & \multicolumn{3}{c}{\emph{horse}} \\ \cline{2-7}
		  &	\emph{head} &  \emph{neck+torso} & \emph{leg}  & \emph{head} &  \emph{neck+torso} & \emph{leg} \\ \hline
			
			PartBB+ObjSeg& 26.77 & 53.79 & 11.18   & 37.32 & 60.35 & 27.47  \\
			PartMask+ObjSeg  & 33.19 & 56.69 & 11.31 & 41.84 & 63.31 & 21.38\\ 
			Compositional model~\cite{wang2015semantic} & 41.55 & 60.98 & 30.98 & 47.21 & 66.74 & 38.18   \\
			 \hline
			DFF &    \cellcolor{myGreen} 40.53 & \cellcolor{myRed} 59.48 & \cellcolor{myBlue} 21.57  &     28.85 & 54.77 & 28.94  \\ 
			DFF-CRF &    45.20 &  58.87 & 24.60  &     31.05 & 53.18 & 28.81  \\
			\hline
		\end{tabularx}
		\caption{Avg. IoU(\%) for three fully supervised methods reported in \cite{wang2015semantic} (see section \ref{sec:baseline} for details) and for our weakly-supervised DFF approach.  As opposed to DFF, previous approaches shown are fully supervised.
			Despite not using hand-crafted features, DFF compares favorably to these approaches, and is not specific to these two image classes. We semi-automatically mapped DFF factors ($k=3$) to their appropriate part labels by examining the heat maps of \emph{only five} images,  out of approximately 140 images. This illustrates the usefulness of DFF co-segmentation for fast semi-automatic labeling. See visualization for \emph{cow} heat maps in Figure \ref{fig:pp-parts}.}
		 \label{tab:wang}
	\end{table}
	
		\begin{table}
			\centering
			\tiny
			
			\begin{tabularx}{\textwidth}
				{Y|l@{\hspace{1pt}}l:l@{\hspace{1pt}}l:l@{\hspace{1pt}}l:l@{\hspace{1pt}}l:l@{\hspace{1pt}}l} 
				\hline
				 $k$  &  \multicolumn{2}{:c}{aeroplane} & \multicolumn{2}{:c}{bird} & \multicolumn{2}{:c}{car} & \multicolumn{2}{:c}{motorbike} & \multicolumn{2}{:c}{cat} \\ \hline
				 \multirow{1}{*}{1} & \emph{aeroplane} & 42 & \emph{bird} & 40 & \emph{car} & 29 & \emph{wheel} & 30 & \emph{eye\slash head\slash neck\slash nose} & 31\\\hline
				 \multirow{2}{*}{2} & \emph{ wheel} & 2 & \emph{beak\slash eye\slash head\slash neck} & 13 & \emph{wheel} & 10 & \emph{wheel} & 38 & \emph{torso} & 24\\
				 & \emph{body\slash stern\slash tail\slash wing} & 49 & \emph{neck\slash torso\slash wing} & 39 & \emph{door\slash roof\slash window} & 22 & \emph{person} & 9 & \emph{eye\slash head\slash neck\slash nose} & 36\\\hline
				 \rowcolor{myBlue}
				 \cellcolor{white}
				 & \emph{wheel} & 2 & \emph{leg} & 2 & \emph{wheel} & 10 & \emph{wheel} & 30 & \emph{eye\slash head\slash neck\slash nose} & 32\\
				 \rowcolor{myRed}
				 \cellcolor{white}
				 & \emph{body\slash stern\slash wing} & 47 & \emph{neck\slash torso\slash wing} & 43 & \emph{door\slash headlight\slash licenseplate} & 24 & \emph{headlight} & 1 & \emph{torso} & 30\\
				 \rowcolor{myGreen}
				 \cellcolor{white}
				 \multirow{-3}{*}{3} & \emph{body\slash tail} & 35 & \emph{beak\slash eye\slash head\slash neck\slash torso} & 30 & \emph{mirror\slash roof\slash window} & 20 & \emph{wheel} & 29 & \emph{ear\slash eye\slash head\slash neck\slash nose} & 38\\\hline
				 \multirow{4}{*}{4} & \emph{wheel} & 1 & \emph{foot\slash leg} & 3 & \emph{wheel} & 9 & \emph{wheel} & 33 & \emph{eye\slash head\slash nose} & 31\\
				 & \emph{body\slash  wheel\slash wing} & 44 & \emph{neck\slash torso\slash wing} & 44 & \emph{headlight\slash licenseplate} & 31 & \emph{person} & 10 & \emph{eye\slash neck\slash nose} & 5\\
				 & \emph{stern\slash tail\slash wing} & 21 & \emph{beak\slash eye\slash head\slash neck\slash torso} & 30 & \emph{front} & 8 & \emph{wheel} & 17 & \emph{ear\slash eye\slash head\slash nose} & 35\\
				 & \emph{body\slash  tail} & 32 & \emph{neck} & 2 & \emph{mirror\slash roof\slash window} & 22 & \emph{background} & 13 & \emph{torso} & 27\\\hline
				 
			\end{tabularx}\\[1ex]
			
			\captionof{table}{IoU of DFF heat maps with PASCAL-Parts segmentation masks. Each DFF factor is autmatically labeled with part labels as in section \ref{sec:icoseg_eval}. Higher values of $k$ allow DFF to localize finer regions across the image set, some of which go beyond the resolution of the ground truth part annotation. Figure \ref{fig:pp-parts} visualizes the results for $k=3$ (row color corrsponds to heat map color).}\label{tab:PP}
		\end{table}

	\begin{figure}[t]
		 \centering
		\setlength{\tabcolsep}{0.8pt}
		\begin{tabularx}{\textwidth}{YY}
			(a) Aeroplane & (b) Bird \\
			\multicolumn{1}{l|}{\begin{minipage}{0.49\textwidth}\includegraphics[width=\textwidth]{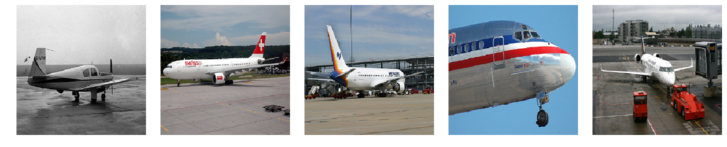}\end{minipage}~~}& \multicolumn{1}{l}{\begin{minipage}{0.49\textwidth}\includegraphics[width=\textwidth]{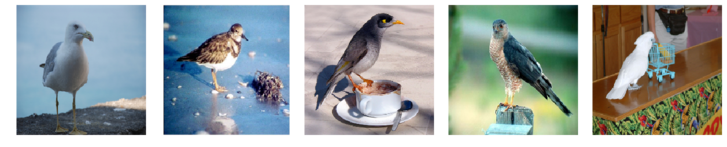}\end{minipage}}  \\
			
			\multicolumn{1}{l|}{\begin{minipage}{0.49\textwidth}\includegraphics[width=\textwidth]{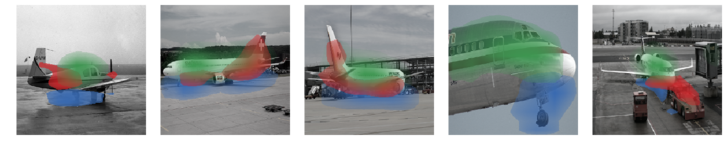}\end{minipage} } & \multicolumn{1}{l}{\begin{minipage}{0.49\textwidth}\includegraphics[width=\textwidth]{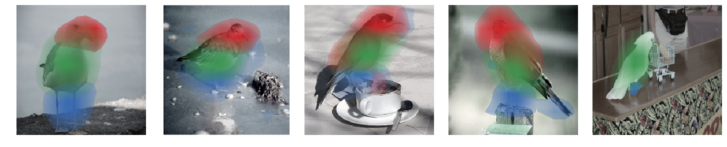}\end{minipage}}
		\end{tabularx}\\[1ex]
		
		\setlength{\tabcolsep}{0.8pt}
		\begin{tabularx}{\textwidth}{YY}
			(c) Car & (d) Cow \\
			\multicolumn{1}{l|}{\begin{minipage}{0.49\textwidth}\includegraphics[width=\textwidth]{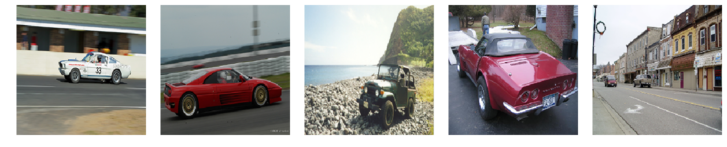}\end{minipage}~~}& \multicolumn{1}{l}{\begin{minipage}{0.49\textwidth}\includegraphics[width=\textwidth]{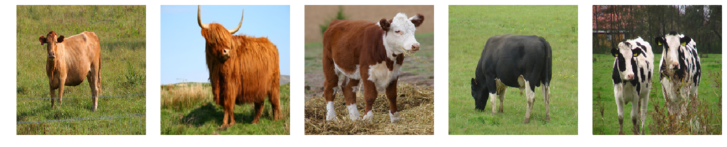}\end{minipage}} \\
			
			\multicolumn{1}{l|}{\begin{minipage}{0.49\textwidth}\includegraphics[width=\textwidth]{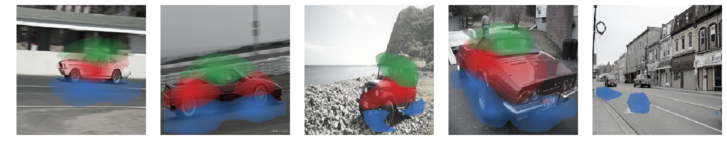}\end{minipage} } & \multicolumn{1}{l}{\begin{minipage}{0.49\textwidth}\includegraphics[width=\textwidth]{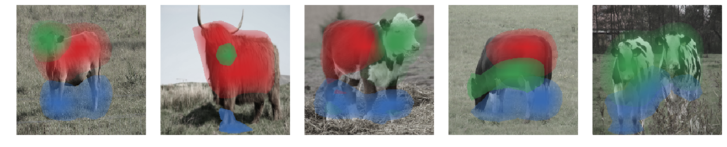}\end{minipage}}
			
		\end{tabularx}\\[1ex]
		
		\setlength{\tabcolsep}{0.8pt}
		\begin{tabularx}{\textwidth}{YY}
			(e) Motorbike & (f) Cat \\
			\multicolumn{1}{l|}{\begin{minipage}{0.49\textwidth}\includegraphics[width=\textwidth]{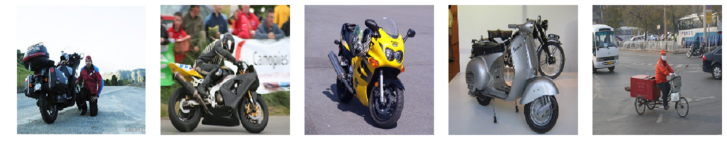}\end{minipage}~~}& \multicolumn{1}{l}{\begin{minipage}{0.49\textwidth}\includegraphics[width=\textwidth]{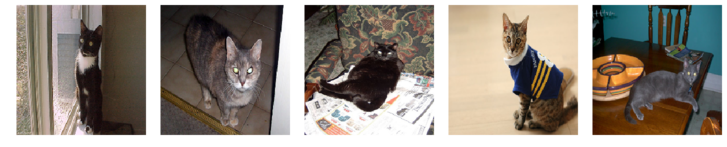}\end{minipage}} \\
			
			\multicolumn{1}{l|}{\begin{minipage}{0.49\textwidth}\includegraphics[width=\textwidth]{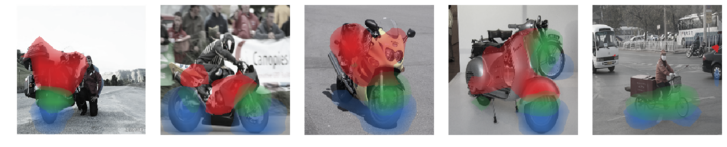}\end{minipage} } & \multicolumn{1}{l}{\begin{minipage}{0.49\textwidth}\includegraphics[width=\textwidth]{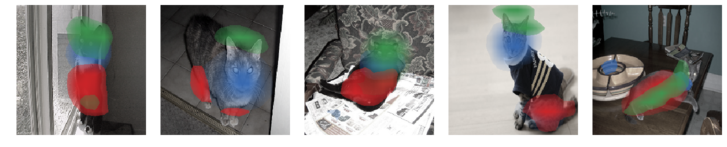}\end{minipage}}  
			
		\end{tabularx}
		
		\caption{Example DFF heat maps for images of six classes from PASCAL-Parts with $k=3$. For each class we show four images that were successfully decomposed into parts, and a failure case on the right. DFF manages to retrieve interpretable decompositions in spite of the great variation in the data. In addition to the DFF factors for \emph{cow} from Table \ref{tab:wang}, here visualized are the factors which appear in Table \ref{tab:PP}, where heat map colors correspond to row colors.}  \label{fig:pp-parts}
	\end{figure}
	
	\subsubsection{Dataset}
	The PASCAL-Part dataset \cite{Chen2014} is an extension of PASCAL VOC 2010 \cite{pascal-voc-2010} which has been further annotated with part-level segmentation masks and bounding boxes. The dataset decomposes 16 object classes into fine grained parts, such as \emph{bird-beak} and \emph{bird-tail} etc. After filtering out images containing objects marked as \emph{difficult} and \emph{truncated}, the final set consists of 16 image sets with 104 to 675 images each.

	\subsubsection{Evaluation}
	
	In Table \ref{tab:wang} we report results for the two classes, \emph{cow} and \emph{horse}, which are also part-segmented by Want et al. as described in section \ref{sec:baseline}. Since their method relies on strong explicit priors w.r.t to part size, hierarchy, and symmetry, and its explicit objective is to perform part-segmentation, their results serve as an upper bound to ours. Nonetheless we compare favorably to their results and even surpass them in one case, despite our method not using any hand-crafted features or supervised training.
	
	For this experiment, our strategy for mapping DFF factors ($k=3$) to their appropriate part labels was with semi-automatic labeling, i.e., we qualitatively examined the heat maps of \emph{only five images}, out of approximately 140 images, and labeled factors as corresponding to the labels shown in Table \ref{tab:wang}.
	
	In Table \ref{tab:PP} we give $IoU$ results for five additional classes from PASCAL-Parts, which have been automatically mapped to parts as in section \ref{sec:icoseg_eval}.
	In Figure \ref{fig:pp-parts} we visualize these DFF heat maps for $k=3$, as well as for \emph{cow} from Table \ref{tab:wang}. When comparing the heat maps against their corresponding $IoU$-scores, several interesting conclusions can be made. For instance, in the case of \emph{motorbike}, the first and third factors for $k=3$ in Table \ref{tab:PP} both seems to correspond with wheel. The visualization in Figure \ref{fig:pp-parts} (e) reveals that these factors in fact sub-segment the wheel into top and bottom, which is beyond the resolution of the ground truth data.
	
	We can see also that while the first factor of the class \emph{aeroplane} (Figure \ref{fig:pp-parts} (a)) consistently localizes airplane wheels, it does not to achieve high $IoU$ due to the coarseness of the heat map. %Alleviating this issue could therefore lead to a significant improvement in localization ability.

	 Returning to Table \ref{tab:PP}, when $k=4$, a factor emerges that localizes instances of the class \emph{person}, which occur in 60\% of motorbike images. This again shows that while most co-localization methods only describe objects that are common across the image set, our DFF approach is able to find distinctions \emph{within} the set of common objects.

	\section{Conclusions}
	In this paper, we have presented Deep Feature Factorization (DFF), a method that is able to locate semantic concepts in individual images and across image sets. We have shown that DFF can reveal interesting structures in CNN feature space, such as hierarchical clusters which correspond to a part-based decomposition at various levels of granularity.
	
	We have also shown that DFF is useful for co-segmentation and co-localization, achieving results on challenging benchmarks which are on par with state-of-the-art methods, and can be used to perform semi-automatic image labeling. Unlike previous approaches, DFF can also perform \emph{part} co-segmentation, making fine distinctions between parts \emph{within} the common object, e.g. matching \emph{head} to \emph{head} and \emph{torso} to \emph{torso}.

	\clearpage
	\bibliographystyle{splncs04}
	\bibliography{egbib}
\end{document}